\documentclass[10pt,twocolumn,letterpaper]{article}

\usepackage[pagenumbers]{cvpr} %

\definecolor{cvprblue}{rgb}{0.21,0.49,0.74}
\usepackage[pagebackref,breaklinks,colorlinks,allcolors=cvprblue]{hyperref}

\usepackage{subcaption}
\usepackage{relsize}
\usepackage{tabularx}
\usepackage{graphicx}
\usepackage{caption}
\usepackage[accsupp]{axessibility}  %
\newsavebox\CBox

\usepackage{xcolor}
\usepackage{overpic}
\definecolor{darkgreen}{rgb}{0.0, 0.5, 0.0}
\definecolor{pink}{rgb}{1,0.75,0.8}
\usepackage{colortbl}

\graphicspath{{images/}}
\usepackage{amsmath}
\usepackage{dsfont}

\usepackage{algpseudocode}
\usepackage[ruled,vlined]{algorithm2e}

\newcommand{\figref}[1]{Fig.~\ref{#1}}
\newcommand{\tabref}[1]{Tab.~\ref{#1}}
\newcommand{\secref}[1]{Sec.~\ref{#1}}

\newcommand{\equref}[1]{Eqn.~(\ref{#1})}
\usepackage{booktabs}
\usepackage{multirow}
\usepackage{overpic}
\usepackage{enumitem}
\usepackage{xcolor}
\usepackage{subcaption}
\usepackage{amsfonts}
\usepackage{amsmath}
\usepackage{float}
\usepackage{booktabs}
\usepackage{wrapfig}

\usepackage[utf8]{inputenc} %
\usepackage[T1]{fontenc}    %
\usepackage{hyperref}       %
\usepackage{url}            %
\usepackage{amsfonts}       %
\usepackage{nicefrac}       %
\usepackage{microtype}      %

\def\paperID{3030} %
\def\confName{CVPR}
\def\confYear{2026}

\title{VGGT-Det: Mining VGGT Internal Priors for Sensor-Geometry-Free \\ Multi-View Indoor 3D Object Detection
}

\author{
Yang Cao$^{1}$\thanks{Equal contribution} \quad
Feize Wu$^{1,3*}$\thanks{Work done during an internship at HKUST} \quad
Dave Zhenyu Chen$^{2}$ \quad
Yingji Zhong$^{1}$ \quad
Lanqing Hong$^{2}$ \quad
Dan Xu$^{1}$\thanks{Corresponding author: danxu@cse.ust.hk}
  \vspace{0.1cm}
  \\
  $^{1}$Hong Kong University of Science and Technology \\
  $^{2}$Huawei \quad
  $^{3}$Sun Yat-Sen University
}

\begin{document}
\maketitle
\begin{abstract}
Current multi-view indoor 3D object detectors rely on sensor geometry that is costly to obtain—i.e., precisely calibrated multi-view camera poses—to fuse multi-view information
into a global scene representation,
limiting deployment in real-world scenes.
We target a more
practical setting: Sensor-Geometry-Free~(\textbf{SG-Free}) multi-view indoor 3D object detection, where there are no
sensor-provided geometric inputs (multi-view poses or depth).
Recent Visual Geometry Grounded Transformer (VGGT) shows that strong 3D cues
can be inferred directly from images. Building on this insight, we present
\textbf{VGGT-Det}, the first framework tailored for SG-Free multi-view indoor 3D
object detection. Rather than merely consuming VGGT predictions,
our method integrates VGGT encoder into a transformer-based pipeline.
To effectively leverage both the semantic and geometric priors
from inside VGGT,
we introduce two novel key components: (i) Attention-Guided
Query Generation (\textbf{AG}): exploits VGGT attention maps as semantic priors to initialize
object queries, improving localization by focusing on object regions while preserving
global spatial structure; (ii) Query-Driven Feature Aggregation (\textbf{QD}): a
learnable \emph{See-Query} interacts with object queries to `see' what they need,
and then dynamically aggregates multi-level geometric features across VGGT layers that progressively
lift 2D features into 3D.
Experiments show that VGGT-Det significantly surpasses the best-performing method in the SG-Free setting by \textbf{4.4} and \textbf{8.6} mAP@0.25 on ScanNet and ARKitScenes,
respectively. Ablation study shows that VGGT’s internally learned semantic and geometric priors can be
effectively leveraged by our AG and QD. 
Source code and pre-trained models are available at the~\href{https://github.com/yangcaoai/VGGT-Det-CVPR2026}{GitHub project page}.

\end{abstract}

\vspace{-0.65cm}

\section{Introduction}
\label{sec:intro}

 Multi-view indoor 3D object detection is a fundamental task  with broad applications in robotics and augmented reality. Current methods~\cite{xu2024mvsdet,xu2023nerf,Shen_2024_CVPR,cao20243dgs,imvotenet} predominantly rely on \emph{sensor-derived geometric inputs}, i.e., precisely calibrated multi-view camera poses or depth. 
Although these methods achieve strong performance, their reliance on expensive and often inaccessible sensor-derived geometric inputs~\cite{cao2023coda,cao2024collaborative} severely limits scalability and real-world deployment.

\begin{figure}[t!]
     \centering
    \begin{overpic}[width=1\columnwidth]
    {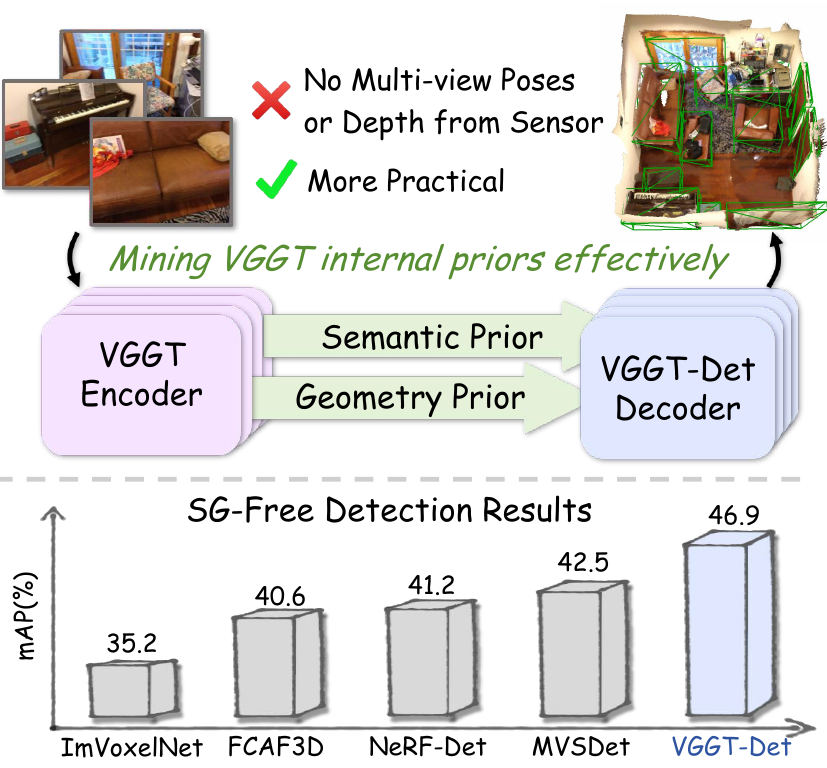}
    \end{overpic}
    \vspace{-0.5cm}  
    \caption{
To achieve and improve Sensor-Geometry-Free~(SG-Free) multi-view indoor 3D object detection, VGGT-Det effectively leverages the internal semantic and geometric priors from VGGT~\cite{wang2025vggt}, rather than merely consuming its predictions. VGGT-Det significantly surpasses competitive methods in the SG-Free setting.
}
\label{fig:teaser}
\vspace{-0.35cm}  
\end{figure}

We instead consider a more practical  setting: performing
indoor 3D detection from multi-view images without sensor-derived geometric inputs. We refer to this as Sensor-Geometry-Free (\textbf{SG-Free}) multi-view indoor 3D object detection. This setting is highly challenging because it eliminates the sensor-derived multi-view camera poses and depth.
Recent advances in feed-forward 3D reconstruction have shown that 3D structure can be inferred directly from unposed 2D images~\cite{wang2024dust3r,leroy2024grounding,yan2024gs,smart2024splatt3r,yang2025fast3r,wang2025vggt}. These developments open new opportunities for SG-Free multi-view indoor 3D object detection. Beyond providing low-cost estimates of scene geometry from unposed images, the feed-forward models also encode strong learned priors that improve 3D reasoning capabilities that are crucial for reliable detection.

In this paper, we present \textbf{VGGT-Det}, a framework for Sensor-Geometry-Free (SG-Free) multi-view indoor 3D object detection, built upon the Visual Geometry Grounded Transformer (VGGT)~\cite{wang2025vggt}, a representative feed-forward 3D reconstruction model. 
As shown in~\figref{fig:teaser}, rather than merely consuming VGGT’s predictions, our method integrates a pretrained VGGT encoder in a transformer-based pipeline. To effectively leverage the learned knowledge from inside VGGT, we conduct an in-depth analysis of its intermediate representations.
We find that the attention maps in the VGGT encoder capture rich semantic information despite the model not being explicitly trained for semantics. Motivated by this interesting observation, we propose Attention-Guided Query Generation (\textbf{AG}), which exploits VGGT attention maps to initialize and steer object queries, improving localization by focusing on semantic regions while preserving global spatial structure.
Furthermore, since VGGT progressively lifts 2D features into 3D across its layers—each encoding distinct levels of geometric abstraction—we introduce Query-Driven Feature Aggregation (\textbf{QD}). This module incorporates a learnable \emph{See-Query} that interacts with object queries to `see' what information is required and then dynamically aggregates multi-level geometric features from VGGT layers, effectively capturing hierarchical representations.
Together, these designs unlock more of VGGT’s potential for SG-Free multi-view indoor 3D detection. Extensive experiments show that VGGT-Det significantly outperforms strong methods in the SG-Free setting, and ablations confirm that the proposed AG and QD effectively leverage the learned priors inside
VGGT.
Our contributions are summarized as follows:
\vspace{-0.2cm}
\begin{itemize}[leftmargin=*]
  \item We introduce a more practical setting—Sensor-Geometry-Free (SG-Free) multi-view indoor 3D detection—which removes the need for sensor-derived geometric inputs (i.e., multi-view camera poses or depth maps). 
  To the best of our knowledge, this is \emph{the first work} to explicitly target this challenging setting.
  Building on this setting, we introduce VGGT-Det, the first transformer-based SG-Free 3D object detection framework. 
  
  \item We propose Attention-Guided Query Generation (AG), which leverages semantic priors encoded in VGGT encoder attention to initialize object queries.
These queries concentrate on object regions while preserving the 
global spatial structure, improving the
localization.

  \item We develop Query-Driven Feature Aggregation (QD), which introduces a learnable \emph{See-Query} that interacts with object queries to `see'  their needs and dynamically aggregates multi-level geometric features accordingly.
  
  \item With the proposed approach, VGGT-Det achieves significant gains, consistently outperforming the best-performing method in the SG-Free setting by \textbf{4.4} and \textbf{8.6} mAP@0.25 points on ScanNet and ARKitScenes, respectively.

\end{itemize}

\vspace{-0.1cm}

\section{Related Work}

\noindent{\textbf{Multi-view Outdoor 3D Object Detection.}
Current outdoor multi-view 3D object detection~\cite{wang2023exploring,xiong2023cape,chen2023viewpoint,feng2023aedet} can be broadly divided into two lines.
The first line projects 3D queries to 2D images
for aggregating visual knowledge.
Inspired by DETR~\cite{carion2020end},
DETR3D~\cite{wang2022detr3d} uses sparse
3D object queries to index multi-view 2D features through camera parameters.
The PETR series~\cite{liu2022petr,liu2023petrv2} introduce a position embedding transformation that encodes 3D coordinate information into multi-view image features. 
The Sparse4D series~\cite{sparse4dv1,sparse4dv2,sparse4dv3} and SparseBEV~\cite{liu2023sparsebev} perform temporally aware detection
by exploring sparse strategies to aggregate multi-frame features.
BEVFormer~\cite{li2024bevformer} introduces a unified BEV representation learned with spatiotemporal transformers.
The second line constructs BEV representations by lifting 2D features
into 3D space.
LSS~\cite{philion2020lift} introduces an end-to-end multi-view
architecture that `lifts' each image to a feature frustum and
`splats' it into a BEV grid to learn a robust scene representation.
BEVDet~\cite{huang2021bevdet} utilizes standard modules but boosts performance via tailored data augmentation and improved NMS.
BEVDepth~\cite{li2023bevdepth} and BEVStereo~\cite{li2023bevstereo} address the depth bottleneck using explicit depth supervision.
Taking a step further from the above methods,
CorrBEV~\cite{xue2025corrbev} tackles occlusion through elegant design choices, such as auxiliary visual and language prototypes.
Our work focuses on indoor detection. Unlike outdoor settings where cameras are rigidly mounted on vehicles, indoor cameras are usually handheld or frequently repositioned. This makes obtaining reliable sensor poses both costly and often inaccessible, motivating
the introduction of the SG-Free setting for indoor scenes.

\begin{figure*}[t!]
     \centering
    \begin{overpic}[width=1\textwidth]
    {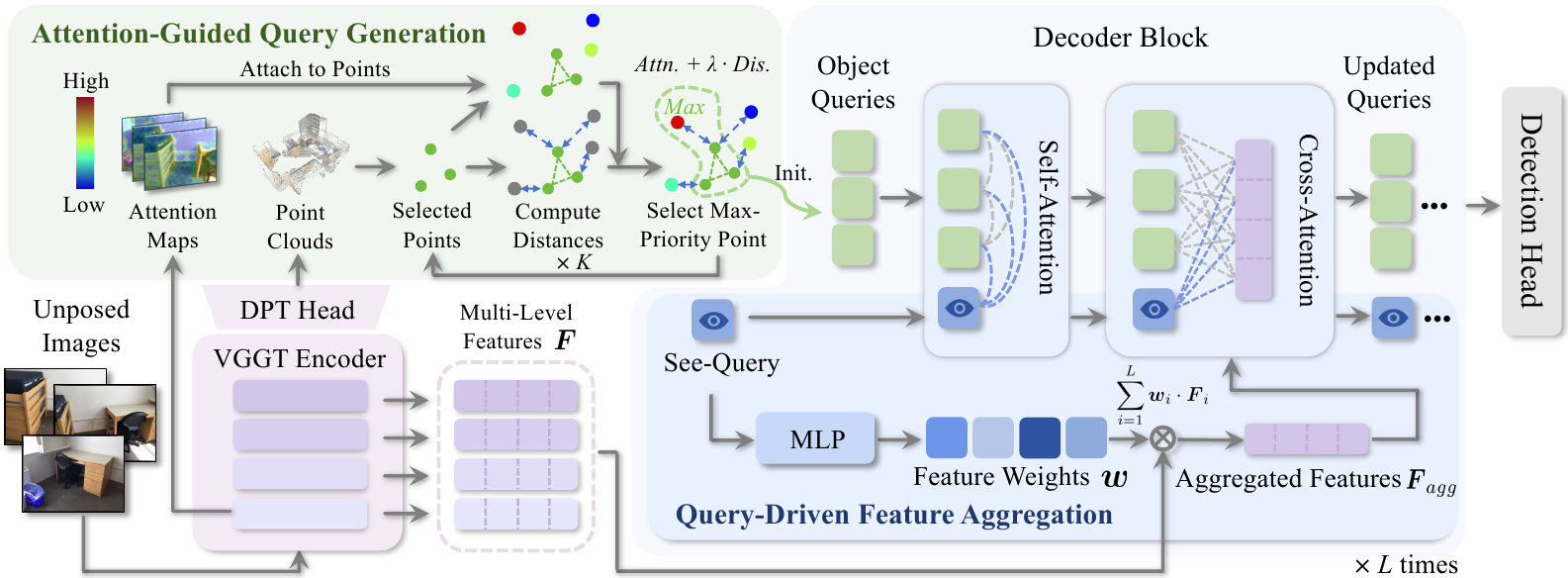}
    \end{overpic}
    \vspace{-0.5cm}  
    \caption{
Overview of the proposed Sensor-Geometry-Free (SG-Free) multi-view indoor 3D object detection framework, \emph{VGGT-Det}. It is built upon the VGGT encoder~\cite{wang2025vggt}, which extracts 3D-aware features from multi-view images. The decoder processes a set of object queries that cross-attend to the extracted features, and iteratively
updates the queries for final detection.  
To effectively leverage both the semantic and
geometric priors from inside VGGT,
we carefully design two key components: \emph{Attention-Guided Query Generation (AG)} and \emph{Query-Driven Feature Aggregation (QD)}. AG utilizes the semantic priors from the VGGT encoder's attention to generate object queries, enabling these queries to focus on object regions while preserving the global spatial structure. Besides, QD introduces a learnable See-Query, which interacts with object queries via self-attention to `see' their needs and dynamically aggregates multi-level geometric features accordingly.
}
    \label{fig:method_overview}
    \vspace{-0.5cm}  
\end{figure*}

\noindent{\textbf{Multi-view Indoor 3D Object Detection.}
Present multi-view indoor 3D object detection methods handle 3D object detection from images with multi-view poses~\cite{xu2024mvsdet,xu2023nerf,Shen_2024_CVPR,cao20243dgs}.
Notably, recent advances in Multi-View Stereo~\cite{yao2018mvsnet,yao2019recurrent}, Neural Radiance Fields (NeRF)~\cite{mildenhall2020nerf}, and 3D Gaussian Splatting~(3DGS)~\cite{kerbl20233d} have enhanced geometry recovery from posed multi-view images, enabling effective 3D detection that leverages 3D clues of voxel~\cite{xu2023nerf,hu2023nerf} and splats~\cite{cao20243dgs}.
ImVoxelNet~\cite{rukhovich2022imvoxelnet} aggregates 2D features from multi-view images into 3D voxel volumes via unprojection in an end-to-end manner.
NeRF-Det~\citep{xu2023nerf} integrates multi-view geometric constraints derived from the NeRF module into the 3D object detection pipeline.
MVSDet~\cite{xu2024mvsdet} replaces NeRF with plane sweep and uses probabilistic sampling with soft weighting to regularize depth, enabling accurate detection.
Existing methods rely on sensor-derived geometric inputs (e.g., multi-view poses or depth), which severely limits their applicability in the real world. To address this, we tackle Sensor-Geometry-Free (SG-Free) multi-view indoor 3D object detection and propose VGGT-Det, which utilizes Attention-Guided Query Generation and Query-Driven Feature Aggregation to effectively mine internal semantic and geometric priors from VGGT rather than merely consuming its predictions.

\noindent{\textbf{Generalizable 3D Reconstruction. }}Given unposed multi-view images, 3D reconstruction~\cite{hartley2003multiple,leroy2024grounding} targets estimating geometry and camera poses of the images. The traditional pipeline for 3D reconstruction involves several sub-tasks, including keypoint detection~\cite{lowe2004distinctive,detone2018superpoint},
matching~\cite{brachmann2019neural,lindenberger2023lightglue,sarlin2020superglue,sun2021loftr}, multi-view stereo~\cite{yao2018mvsnet,yao2019recurrent,wang2021patchmatchnet,zhang2020visibility}, etc. 
Recently, DUSt3R~\cite{wang2024dust3r} revolutionizes the paradigm and uses a single network to directly estimate the geometry of a target scene. 
MASt3R~\cite{leroy2024grounding} inherits the paradigm with an auxiliary prediction head for the matching task. 
Both DUSt3R and MASt3R are limited to pair-wise inputs, which restricts context reasoning and leads to repeated network forward passes and time-consuming global coordinate alignment. 
Fast3R~\cite{yang2025fast3r} improves performance by enabling a single forward pass for long sequences and eliminating the need for coordinate alignment. 
In addition to point maps, VGGT~\cite{wang2025vggt} predicts camera poses and other 3D-relevant attributes. 
In this paper, we present VGGT-Det, which effectively leverages priors within VGGT to advance SG-Free multi-view indoor 3D object detection.

\section{Method}

\subsection{Overview}
An overview of our proposed sensor-geometry-free 3D detection framework is presented in \figref{fig:method_overview}.
VGGT-Det takes multi-view images
as inputs and outputs 3D detection results, which eliminates
sensor-provided geometric inputs (i.e., multi-view poses and depth).
In the following, we will first introduce
our basic backbone in~\secref{basic-pipeline}, 
which is a transformer-based pipeline integrating the VGGT encoder~\cite{wang2025vggt}. 
To effectively leverage both the semantic and geometric
priors from inside VGGT, we introduce two novel key components:
(i) Attention-Guided Query Generation (AG) is proposed
to effectively utilize the semantic prior from
VGGT encoder attention, detailed in~\secref{ag-query}.
(ii) Query-Driven Feature Aggregation (QD) is designed
to adaptively aggregates multi-level geometric
features according to the needs of object queries, 
which is introduced
in~\secref{qd-feature}.

\subsection{Basic Backbone} \label{basic-pipeline}

We design an encoder-decoder transformer~\cite{vaswani2017attention} architecture as our backbone. Given a set of multi-view images $\{I_1, I_2, \dots, I_V\}$, we employ the VGGT encoder~\cite{wang2025vggt} as the feature extractor to extract 3D-aware features. For each view, the VGGT encoder outputs a sequence of tokens, where each token corresponds to a spatial region or a patch of the image. The extracted tokens for all views are denoted as $\{\mathbf{T}_1, \mathbf{T}_2, \dots, \mathbf{T}_V\}$, where $\mathbf{T}_i \in \mathbb{R}^{M \times C}$ is the token sequence for the $i$-th view, with $M$ tokens per view and $C$ as the token dimension. These tokens are concatenated along the token dimension to form a unified token representation:

\vspace{-0.2cm}
\begin{equation}
\mathbf{T}_{\text{concat}} = [\mathbf{T}_1; \mathbf{T}_2; \dots; \mathbf{T}_V] \in \mathbb{R}^{(V \cdot M) \times C}.
\end{equation}

\begin{figure}[t!]
    \centering
    \includegraphics[width=1\linewidth]{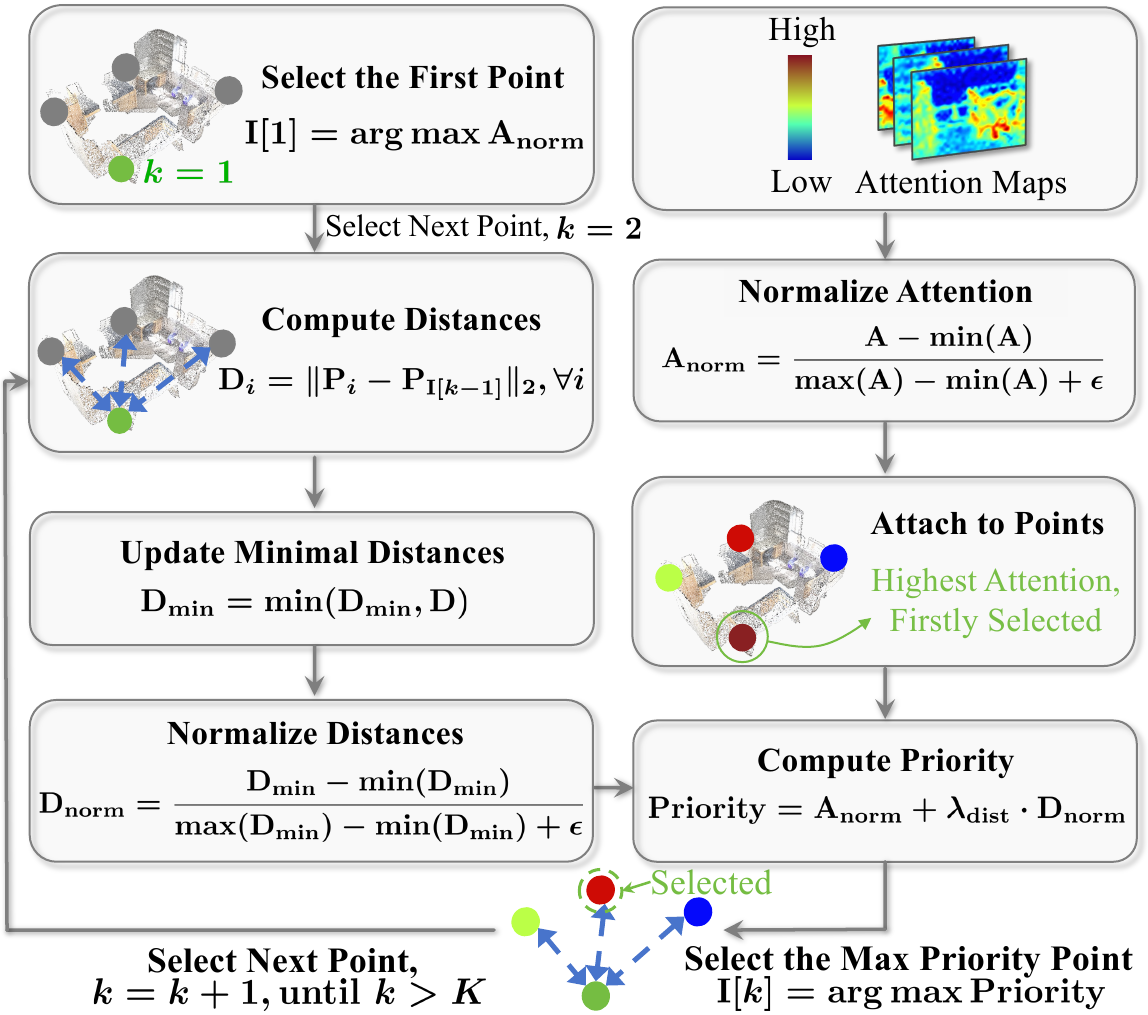}
     \vspace{-0.6cm}
    \caption{Computation flow of Attention-Guided Query Generation.}
    \label{fig:AG_method}
    \vspace{-0.5cm}
\end{figure}

Inspired by 3DETR~\cite{misra2021end}, we adopt non-parametric query sampling to initialize object queries. Specifically, we apply farthest point sampling (FPS) to the point clouds $\mathbf{P}_{\text{pred}}$ predicted by VGGT. FPS selects a subset of $K$ points $\{\mathbf{p}_1, \mathbf{p}_2, \dots, \mathbf{p}_K\} \subset \mathbf{P}_{\text{pred}}$, where $K$ is the number of object queries. These sampled points are encoded with positional embeddings $\mathbf{E}_{\text{pos}} \in \mathbb{R}^{K \times C}$ to initialize object queries:

\vspace{-0.2cm}
\begin{equation}
\mathbf{Q}_0 = \text{Embed}(\{\mathbf{p}_1, \mathbf{p}_2, \dots, \mathbf{p}_K\}) + \mathbf{E}_{\text{pos}}.
\end{equation}

The object queries $\mathbf{Q}_0$ are then passed through a series of $L$ transformer decoder layers. In each decoder layer, the queries first undergo query-to-query self-attention to exchange information among themselves as:

\vspace{-0.2cm}
\begin{equation}
\mathbf{Q}_l^{\text{self}} = \text{SelfAttention}(\mathbf{Q}_{l-1}),
\end{equation}
where $\mathbf{Q}_{l-1}$ is the input to the $l$-th decoder layer, and $\mathbf{Q}_l^{\text{self}}$ is the output after self-attention. Next, the queries attend to the unified token representation $\mathbf{T}_{\text{concat}}$ via cross-attention:

\vspace{-0.2cm}
\begin{equation}
\mathbf{Q}_l = \text{CrossAttention}(\mathbf{Q}_l^{\text{self}}, \mathbf{T}_{\text{concat}}).
\end{equation}

After passing through all $L$ decoder layers, the updated object queries $\mathbf{Q}_L \in \mathbb{R}^{K \times C}$ are fed into a detection head to generate object-level predictions, including class labels $\hat{\mathbf{c}} \in \mathbb{R}^K$ and bounding boxes $\hat{\mathbf{b}} \in \mathbb{R}^{K \times 7}$:

\vspace{-0.2cm}
\begin{equation}
\{\hat{\mathbf{c}}, \hat{\mathbf{b}}\} = \text{DetectionHead}(\mathbf{Q}_L).
\end{equation}

By concatenating tokens across views, the model enables object queries to directly interact with the multi-view token representation. This pipeline effectively leverages the 3D-aware features extracted by the VGGT encoder and dynamically refines object queries through the transformer decoder, enabling the sensor-geometry-free 3D object detection.

\subsection{Attention-Guided Query Generation} \label{ag-query}

\begin{figure*}[t!]
     \centering
    \begin{overpic}[width=1\textwidth]
    {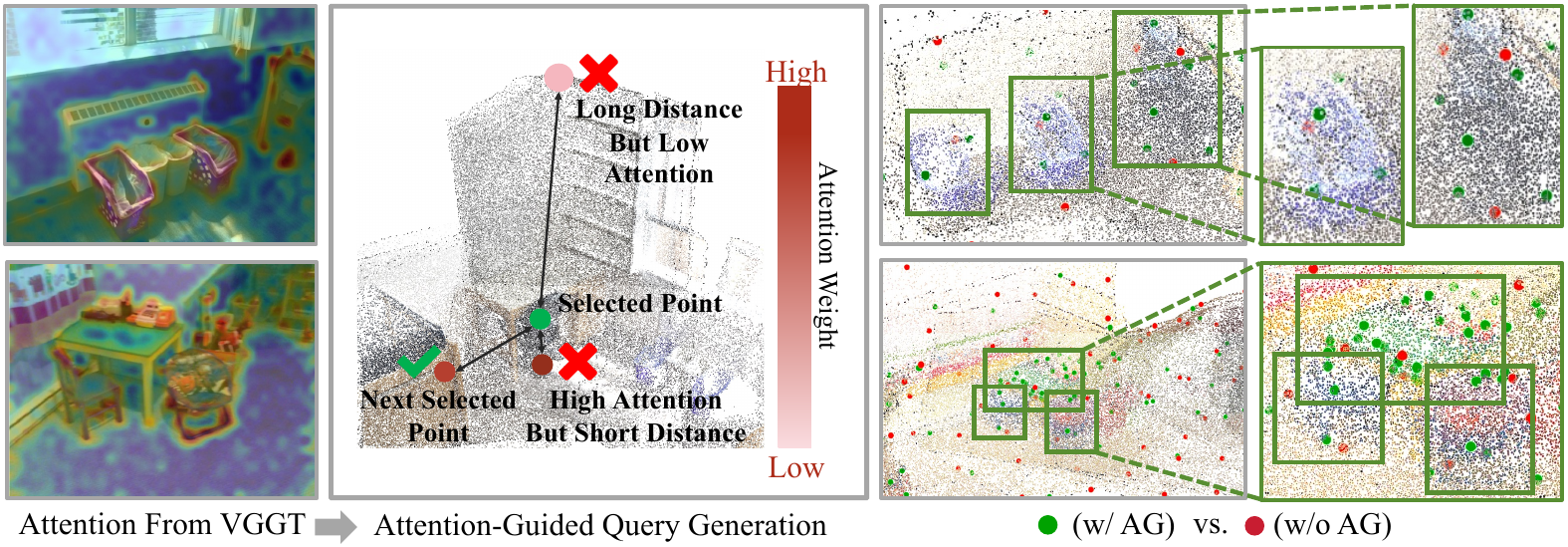}
    \end{overpic}
    \vspace{-0.7cm}
    \caption{
The proposed Attention-Guided Query Generation (AG) is inspired by the interesting observation:
attention maps from the VGGT encoder~\cite{wang2025vggt} exhibit a strong correlation with semantic content, even though VGGT is not explicitly trained for semantic tasks.
For example, in the left column, object regions tend to receive higher attention weights.
In the middle column,
AG samples from VGGT-predicted point clouds under the guidance of attention weights and point distribution information.  
In the right column,
compared to farthest point sampling without guidance~(red points), the points sampled by AG~(green points) are more concentrated in object regions~(labeled by green boxes),
resulting in more green points than red points in those areas. For clarity, we recommend viewing the figure in color and zooming in.
    }
    \label{fig:atten_guided}
    \vspace{-0.4cm}
\end{figure*}

Since the point clouds predicted by VGGT~\cite{wang2025vggt} are dense reconstruction of the scene, it does not distinguish between object and background regions. Therefore, simply applying the FPS to uniformly select queries from the predicted point clouds may result in a significant number of object queries falling in background regions rather than actual objects. 
Such misplacement of queries fundamentally compromises detection efficacy, as background-located queries provide minimal useful information for detection. 
Interestingly, we observed that the attention in the VGGT encoder inherently captures rich semantic information, as shown in the left column of~\figref{fig:atten_guided}, even though VGGT is not trained on semantic tasks.
The observation motivates us to propose the Attention-Guided Query Generation (AG) that leverages the attention maps from the VGGT encoder to encourage object queries to locate at semantic object regions. The proposed AG not only leverages the semantic guidance from the VGGT encoder, but also considers the spatial diversity to make the sampled points fully cover the entire 3D space.  

Formally, the point clouds reconstructed by VGGT are denoted as $\mathbf{P} \in \mathbb{R}^{N \times 3}$, where $N$ is the number of points, and each point is represented by its 3D coordinates. We denote the attention weights from the VGGT encoder as $\mathbf{A} \in \mathbb{R}^{N}$, where each weight indicates the semantic importance of a point. The goal is to sample $K$ query points $\{\mathbf{P}_1, \mathbf{P}_2, \dots, \mathbf{P}_K\}$ from $\mathbf{P}$, guided by both semantic information from the attention weights and spatial distribution.
To ensure numerical stability and compatibility with distance-based features, the attention weights are normalized:

\vspace{-0.2cm}
\begin{equation}
\mathbf{A}_{\text{norm}} =
\frac{
\mathbf{A} - \min(\mathbf{A}
)}{
\max (\mathbf{A}) - \min (\mathbf{A}) + \epsilon
},
\end{equation}
where $\epsilon$ is a small constant for numerical stability.
$\mathbf{A}_{\text{norm}} \in \mathbb{R}^{N}$.
The index of the first query point $\mathbf{I}[1]$ is selected as the point with the highest attention score:

\vspace{-0.1cm}
\begin{equation}
\mathbf{I}[1] = \arg\max \mathbf{A}_{\text{norm}}.
\end{equation}

Following points are selected iteratively by a fused priority that balances semantic attention and spatial dispersion:

\vspace{-0.2cm}
\begin{equation}
\text{Priority} = \mathbf{A}_{\text{norm}} + \lambda_{\text{dist}} \cdot \mathbf{D}_{\text{norm}},
\label{equ:lambda}
\end{equation}
where $\lambda_{\text{dist}} \in [0, 1]$ is a trade-off coefficient that controls the relative influence of spatial dispersion, and $\mathbf{D}_{\text{norm}}$ is the normalized minimum Euclidean distance between point $\mathbf{P}$ and the already-sampled set $\{\mathbf{P}_{\mathbf{I}[1]}, \dots, \mathbf{P}_{\mathbf{I}[k-1]}\}$. Specifically, we let $\mathbf{D}_{\text{min}}$ denote the minimum Euclidean distance from $\mathbf{P}$ to all the previously selected points:

\begin{equation}
\mathbf{D}_{\text{min}} = \min_{j \in \{1, \dots, k-1\}} \lVert \mathbf{P} - \mathbf{P}_{\mathbf{I}[j]} \rVert_2,
\end{equation}
where $ \mathbf{D}_{\text{min}} \in \mathbb{R}^{N}$.
The normalized distance is defined as:

\begin{equation}
\mathbf{D}_{\text{norm}} =
\frac{
\mathbf{D}_{\text{min}} - \min (\mathbf{D}_{\text{min}})
}{
\max (\mathbf{D}_{\text{min}}) - \min (\mathbf{D}_{\text{min}}) + \epsilon
},
\end{equation}
where $ \mathbf{D}_{\text{norm}} \in \mathbb{R}^{N}$. At each iteration, the point with the highest priority is subsequently  selected:

\vspace{-0.2cm}
\begin{equation}
\mathbf{I}[k] = \arg\max_{i \notin \mathcal{S}} \text{Priority}, %
\end{equation}
where $\mathcal{S} = \{\mathbf{I}[1], \mathbf{I}[2],..., \mathbf{I}[k-1] \}$ represents the index set of samples selected in previous iterations.
In this step, as illustrated in the middle column~\figref{fig:atten_guided}, points with high attention scores and greater distances are selected. This design encourages object queries to focus on semantically meaningful object regions while preserving spatial diversity, ensuring coverage of the entire 3D space.
The computation flow is shown in~\figref{fig:AG_method}, which presents a step-by-step implementation of our proposed Attention-Guided Query Generation (AG). 
As shown in the right column of~\figref{fig:atten_guided}, while FPS tends to distribute sampling points indiscriminately across both foreground and background regions, AG strategically concentrates sampling queries on object regions by leveraging
the internal semantic priors from VGGT, facilitating more effective 3D object detection.

\begin{table*}[t]
\centering
\begingroup
{\footnotesize\relsize{+0.5} 
\setlength{\tabcolsep}{3pt}     %
\renewcommand{\arraystretch}{0.94} %

\begin{tabularx}{\textwidth}{l|*{10}{>{\centering\arraybackslash}X}}
\toprule
\textbf{Methods} & cab & bed & chair & sofa & tabl & door & wind & bkshf & pic & cntr \\
\midrule
ImVoxelNet \citep{rukhovich2022imvoxelnet} & 20.2 & 76.3 & 45.8 & 60.5 & 36.0 & 16.3 & 09.6 & 17.9 & 00.3 & 23.3 \\
FCAF3D \citep{rukhovich2022fcaf3d} & 28.1 & 81.0 & 47.2 & 70.7 & 40.9 & 17.1 & 13.2 & 22.0 & 00.7 & 45.6 \\
NeRF-Det \citep{xu2023nerf} & 24.9 & 85.3 & 54.4 & 64.3 & 42.2 & 24.3 & 12.7 & 31.0 & 01.0 & 39.3 \\
MVSDet \citep{xu2024mvsdet} & 28.0 & 80.6 & 55.5 & 65.3 & 43.2 & 25.2 & 09.9 & 29.4 & 01.1 & 43.4 \\
\rowcolor[gray]{.9} VGGT-Det~(Basic backbone)  & 20.1 & 85.5 & 48.4 & 78.5 & 37.0 & 13.2 & 09.4 & 22.1 & 00.8 & 40.4 \\
\rowcolor[gray]{.9} VGGT-Det~(Basic backbone+AG) & 25.9 & 84.7 & 53.7 & 73.8 & 46.3 & 17.6 & 13.7 & 26.4 & 02.2 & 40.5 \\
\rowcolor[gray]{.9} VGGT-Det~(Basic backbone+AG+QD) & 27.8 & 82.0 & 61.9 & 74.0 & 50.3 & 18.3 & 17.6 & 36.9 & 02.6 & 40.3 \\
\midrule
\textbf{Methods} & desk & curt & fridg & showr & toil & sink & bath & ofurn & \multicolumn{2}{c}{\textbf{mAP@0.25}} \\
\midrule
ImVoxelNet \citep{rukhovich2022imvoxelnet} & 53.1 & 08.1 & 29.4 & 35.0 & 83.2 &  43.1 & 61.3 & 14.4 & \multicolumn{2}{c}{35.2} \\
FCAF3D \citep{rukhovich2022fcaf3d} & 65.4 & 08.3 & 32.3 & 39.1 & 83.6 &  46.9 & 72.7 & 15.6 & \multicolumn{2}{c}{40.6} \\
NeRF-Det \citep{xu2023nerf} & 62.6 & 17.3 & 41.4 & 32.8 & 88.4 & 39.7 & 58.7 & 21.0 & \multicolumn{2}{c}{41.2} \\
MVSDet \citep{xu2024mvsdet} & 63.8 & 20.4 & 39.2 & 36.7 & 89.7 & 44.2 & 68.4 & 20.2 & \multicolumn{2}{c}{42.5} \\
\rowcolor[gray]{.9} VGGT-Det~(Basic backbone) & 68.3 & 19.3 & 35.5 & 41.4 & 89.5 & 39.7 & 76.8 & 20.0 & \multicolumn{2}{c}{41.4} \\
\rowcolor[gray]{.9} VGGT-Det~(Basic backbone+AG) & 66.5 & 18.7 & 41.5 & 45.2 & 84.6 & 45.3 & 84.0 & 24.3 & \multicolumn{2}{c}{44.2} \\
\rowcolor[gray]{.9} VGGT-Det~(Basic backbone+AG+QD) & 66.9 & 31.4 & 49.0 & 32.9 & 90.3 & 50.8 & 83.2 & 27.6 & \multicolumn{2}{c}{\textbf{46.9~\footnotesize{\textcolor{darkgreen}{(+4.4)}}}} \\
\bottomrule
\end{tabularx}
}
\endgroup

\begingroup
\captionsetup{aboveskip=2pt,belowskip=0pt}
\caption{Comparison of mAP@0.25 across different methods on ScanNet. To achieve Sensor-Geometry-Free (SG-Free) setting and ensure a fair comparison, FCAF3D is retrained with point clouds predicted by VGGT. ImVoxelNet, NeRF-Det and MVSDet are retrained with multi-view camera poses predicted by VGGT. Our method significantly outperforms the competitive methods.}
\label{tab:main_result_0.25}
\endgroup
\vspace{-0.3cm}
\end{table*}

\subsection{Query-Driven Feature Aggregation} \label{qd-feature}

Recognizing that the VGGT encoder~\cite{wang2025vggt} progressively transforms 2D image features into 3D representations across its layers, with each stage capturing distinct levels of geometric information,
we propose a Query-Driven Feature Aggregation (QD) strategy.
This approach introduces a learnable \emph{See-Query},
which interacts with object queries to `see' their requirements and dynamically
aggregates multi-level features accordingly.
Formally, let the VGGT encoder produce a set of feature maps from $L$ intermediate layers, denoted as $\{\mathbf{F}_1, \mathbf{F}_2, \dots, \mathbf{F}_L\}$, where $\mathbf{F}_i \in \mathbb{R}^{N \times C}$ is the feature map from the $i$-th layer, with $N$ spatial elements and $C$ feature dimensions. To aggregate these multi-level geometric features, we introduce a learnable \emph{See-Query token} $\mathbf{q}_{\text{see}} \in \mathbb{R}^C$, shared across all decoder layers. Firstly, we present how the See-Query aggregates multi-level features. The See-Query token is first transformed by a multi-layer perceptron (MLP) followed by a softmax normalization, producing attention weights $\mathbf{w} \in \mathbb{R}^L$ over the $L$ feature maps:
\vspace{-0.2cm}
\begin{equation}
\mathbf{w} = \textrm{Softmax}\left(\text{MLP}(\mathbf{q}_{\text{see}})\right), \ \mathbf{w}_i \geq 0, \ \sum_{i=1}^L \mathbf{w}_i = 1.
\label{equ:softmax}
\end{equation}
The aggregated feature $\mathbf{F}_{\text{agg}}$ is then computed as a weighted sum of the multi-level geometric encoded features:
\vspace{-0.2cm}
\begin{equation}
\mathbf{F}_{\text{agg}} = \sum_{i=1}^L \mathbf{w}_i \cdot \mathbf{F}_i,
\vspace{-0.1cm}
\end{equation}
where $\mathbf{F}_{\text{agg}} \in \mathbb{R}^{N \times C}$ serves as the key and value inputs to the decoder’s cross-attention block. 
Then, we present how the See-Query interacts with
object queries and is updated.
The See-Query token $\mathbf{q}_{\text{see}}$ plays a dual role in the decoder. First, it is concatenated with the object queries $\{\mathbf{q}_1, \mathbf{q}_2, \dots, \mathbf{q}_K\}$, forming a unified query set  jointly  for decoding:
\vspace{-0.2cm}
\begin{equation}
\mathbf{Q}_{\text{input}} = [\mathbf{q}_{\text{see}}, \mathbf{q}_1, \mathbf{q}_2, \dots, \mathbf{q}_K] \in \mathbb{R}^{(K+1) \times C}.
\end{equation}

This unified query set undergoes a self-attention operation to exchange information among the queries:
\vspace{-0.2cm}
\begin{equation}
\mathbf{Q}_{\text{self}} = \text{SelfAttention}(\mathbf{Q}_{\text{input}}),
\end{equation}
where $\mathbf{Q}_{\text{self}} \in \mathbb{R}^{(K+1) \times C}$. Through this step, the See-Query interacts with the object queries to `see' what they need, and then accordingly enriches them with relevant global multi-scale context
from $\mathbf{F}_{\text{agg}}$. 
Specifically, the See-Query and object queries participate in the cross-attention operation, attending to the aggregated geometric encoder features $\mathbf{F}_{\text{agg}}$:

\vspace{-0.2cm}
\begin{equation}
\mathbf{Q}_{\text{cross}} = \text{CrossAttention}(\mathbf{Q}_{\text{self}}, \mathbf{F}_{\text{agg}}),
\end{equation}
where $\mathbf{Q}_{\text{cross}} \in \mathbb{R}^{(K+1) \times C}$. This generates updated representations for the See-Query and object queries based on the hierarchical encoder features.
After the cross-attention operation, the updated See-Query token $\mathbf{q}_{\text{see}}^{(l)}$ and the object queries are passed to the next decoder layer, where the process repeats starting from \equref{equ:softmax}.
The iterative refinement of the See-Query enables dynamic, context-aware guidance of multi-level geometric feature aggregation across decoder layers, facilitating fine-grained 3D detection.

\vspace{-0.1cm}
\section{Experiments}

\subsection{Experimental Setup}

\noindent{\textbf{Datasets and Settings.}}
We evaluate our method on the widely-used ScanNet~\cite{dai2017scannet} and ARKitScenes~\citep{arkitscenes}. They cover diverse indoor environments like homes, offices, and classrooms. ScanNet includes 18 standard object categories.
ARKitScenes consists of 17 categories. For evaluation, we report mAP@0.25, where the Mean Average Precision (mAP) is computed to assess the model's detection performance across all categories comprehensively.

\noindent{\textbf{Implementation Details.}}
The pretrained VGGT encoder is frozen during training to preserve its original capabilities. In our experiments, we use 256 object queries. The model is optimized using the AdamW optimizer~\cite{loshchilov2017decoupled} with an initial learning rate of $2.5 \times 10^{-4}$ and a weight decay of $1 \times 10^{-4}$. To ensure stable training, gradient clipping is applied with a maximum norm of 35 and a norm type of 2. For learning rate scheduling, we adopt a cosine annealing strategy~\cite{loshchilov2017sgdr}, where the learning rate gradually decays to $1 \times 10^{-6}$, facilitating smooth convergence and mitigating instability caused by abrupt changes in the learning rate. The training loss setting follows 3DETR~\cite{misra2021end}.
In our basic backbone, we employ a naive encoded feature aggregation method, where the encoded features from the 4th, 11th, 17th, and 23rd layers (following the VGGT architecture) are sequentially queried in the decoder. These features, arranged from shallow to deep, are queried step by step to progressively refine the object representations. 
We train our models on 8×H800 GPUs, requiring approximately two days to complete.

\subsection{Method Analysis}
In this section, we present an ablation study to thoroughly analyze the contributions of our proposed designs. We progressively incorporate each design into our basic backbone to evaluate their individual effectiveness.

\noindent{\textbf{Effect of Attention-Guided Query Generation}}. As shown in~\tabref{subtab:components}, our Attention-Guided Query Generation~(the second row) significantly improves the performance of the basic backbone~(the first row), achieving a \textbf{+2.8} point gain. This demonstrates the superiority of our Attention-Guided Query Generation compared to the FPS Sampling Generation. The gains suggest that our Attention-Guided Query Generation effectively leverages the internal semantic priors of VGGT, guiding the object queries to focus on object regions while preserving the global spatial structure of the scene.

\begin{table*}[!h]
\centering
\begingroup
\captionsetup[sub]{labelfont={small}}
{\footnotesize\relsize{+0.8} 
\begin{tabular}{@{}cccc@{}}
\subcaptionbox{\label{subtab:components}}{
\begin{tabular}[t]{l|c}
\toprule
\textbf{Methods} & \textbf{mAP} \\
\midrule
BB & 41.4 \\
BB+AG & 44.2 \\
\rowcolor[gray]{.9} BB+AG+QD & \textbf{46.9} \\
\bottomrule
\end{tabular}}
&
\subcaptionbox{\label{subtab:frames}}{
\begin{tabular}[t]{l|c}
\toprule
\textbf{Number} & \textbf{mAP} \\
\midrule
40 & 45.3 \\
60 & 46.2 \\
\rowcolor[gray]{.9} 80 & \textbf{46.9} \\
\bottomrule
\end{tabular}}
&
\subcaptionbox{\label{subtab:aggregation}}{
\begin{tabular}[t]{l|c}
\toprule
\textbf{Methods} & \textbf{mAP} \\
\midrule
Vanilla-v1 & 37.5 \\
Vanilla-v2 & 44.2 \\
\rowcolor[gray]{.9} QD (Ours) & \textbf{46.9} \\
\bottomrule
\end{tabular}}
&
\subcaptionbox{\label{subtab:efficiency}}{
\begin{tabular}[t]{l|c|c|c}
\toprule
\textbf{Methods} & \textbf{Time (s)} & \textbf{Mem (GB)} & \textbf{mAP} \\
\midrule
VGGT & 0.68 & 11.71 & - \\
w/ MVSDet~\cite{xu2024mvsdet} & 0.21 & 13.81 & 39.7 \\
\rowcolor[gray]{.9} w/ Ours & 0.23 & 3.57 & \textbf{45.3} \\
\bottomrule
\end{tabular}}
\\
\end{tabular}}
\endgroup
\vspace{-0.4cm}
\caption{Ablation studies: (a) Effectiveness of the proposed designs. `BB' denotes our basic backbone; (b) Performance across different numbers of input frames; (c) Comparison of different feature aggregation methods; (d) Time and memory analysis. VGGT computation is in the first row, while the additional time and memory consumption of each method are shown in subsequent rows.}
\vspace{-0.5cm}
\label{tab:abla}
\end{table*}

\begin{figure}[t!]
     \centering
    \begin{overpic}[width=1\columnwidth]
    {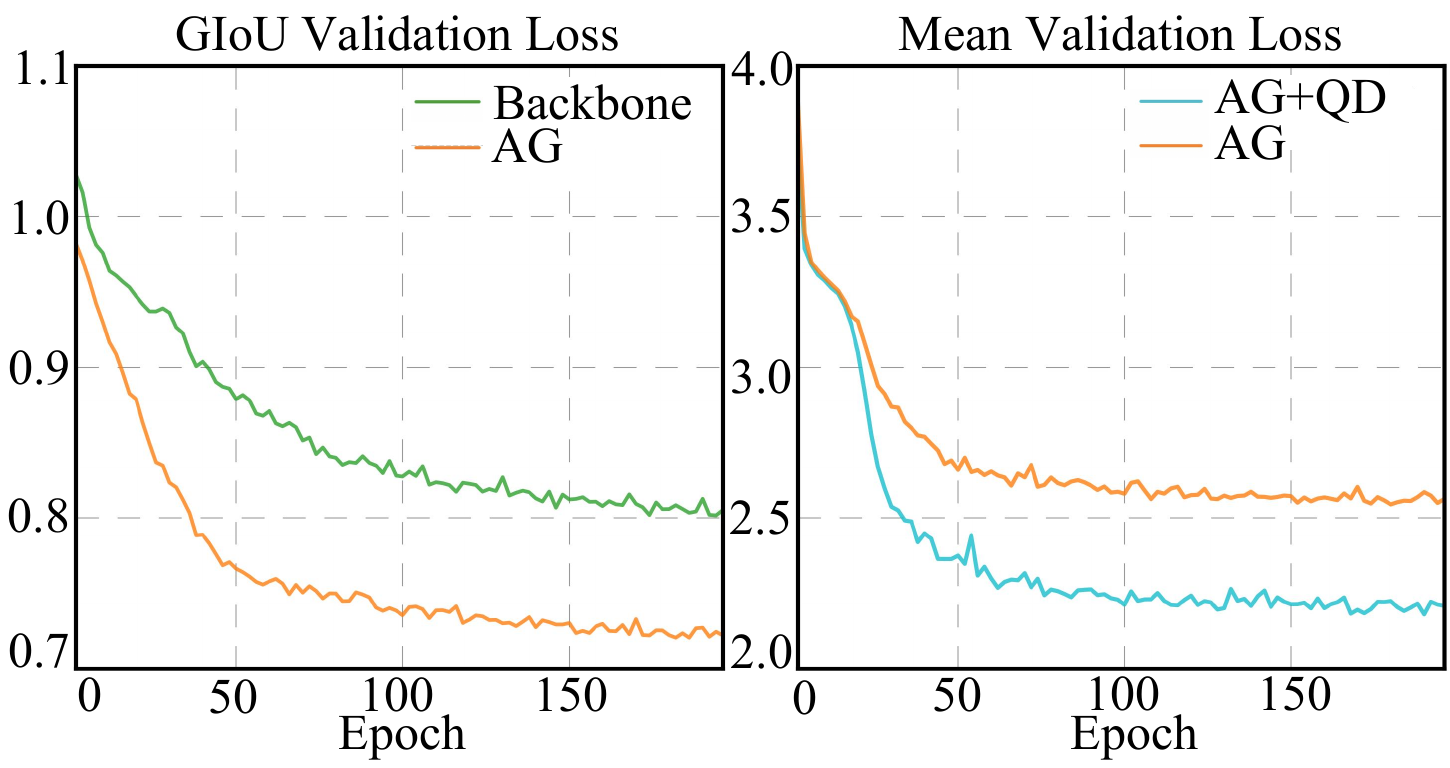}
    \end{overpic}
    \vspace{-0.7cm}  
    \caption{Visualization of validation losses. In the left subfigure, after applying AG, the GIoU loss is significantly lower than  that of the baseline backbone, indicating that AG effectively improves object localization during training.
In the right subfigure, as See-Query progressively learns, within a few epochs, to interact with object queries and to aggregate encoded geometric features effectively, the loss for `AG+QD' becomes significantly lower than that for `AG', highlighting the effectiveness of the proposed QD strategy. }
    \label{fig:loss_curve}
    \vspace{-0.6cm}
\end{figure}

\noindent{\textbf{Effect of Query-Driven Feature Aggregation}}. 
Considering both the first and second rows in ~\tabref{subtab:components}, which utilize the naive encoded feature aggregation strategy, the encoded features from the 4th, 11th, 17th, and 23rd layers (following the default setting of VGGT~\cite{wang2025vggt}) are sequentially queried in the decoder. This straightforward strategy aggregates multi-level geometric features in a simple manner.
In contrast, our proposed Query-Driven Feature Aggregation adaptively aggregates multi-level
geometric features through interactions with object queries. This method achieves a notable improvement, increasing mAP@0.25 by \textbf{2.7} points, showcasing the effectiveness of the See-Query mechanism. By interacting with object queries across different decoder layers, the See-Query learns to identify which levels of encoded features are most relevant at each stage, leading to precise aggregation and significant performance gains.

\begin{figure*}[t!]
     \centering
    \begin{overpic}[width=1\textwidth]
    {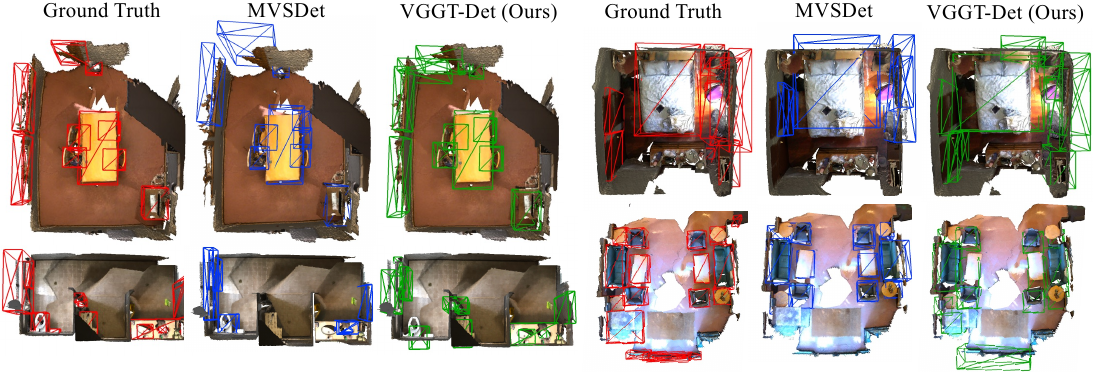}
    \end{overpic}
    \vspace{-0.8cm}     
    \caption{Qualitative comparison.
Compared with MVSDet~\cite{xu2024mvsdet},
our VGGT-Det detects more objects
with higher precision.
}
\vspace{-0.1cm}
\label{fig:demo_comparison}
\end{figure*}

\noindent{\textbf{Validating the Motivation Through Loss Dynamics}}. 
To deeply validate the motivation behind our technical contributions,
we analyze the validation
loss dynamics during training.
Specifically, we evaluate the effects of Attention-Guided Query Generation (`AG')
and its combination with Query-Driven Feature Aggregation (`AG+QD').
As shown in the left subfigure of~\figref{fig:loss_curve}, the GIoU loss with `AG'
is consistently lower than that of the baseline backbone. This supports our motivation
that attention maps effectively help object queries focus on semantic regions,
thereby improving object localization accuracy.
In the right subfigure, we further examine how Query-Driven Feature Aggregation (`AG+QD') complements `AG'.
During the initial epochs, the losses of `AG' and `AG+QD' are similar,
as the See-Query module is still learning to interact with object queries. 
However, after a few epochs, the loss for `AG+QD' becomes significantly lower than that for `AG'.
This demonstrates that the See-Query module learns to effectively aggregate encoded geometric features.
These validation loss dynamics provide strong evidence validating our motivation and confirm
the effectiveness of both `AG' and `AG+QD' in improving the model's training.

\noindent{\textbf{Time and Memory Analysis}}. 
We conducted the time and memory analysis in~\tabref{subtab:efficiency}. For fairness, all experiments used 40 input frames on a 1×H800 GPU, with evaluation performed on ScanNet. The table breaks down the time required to process a 3D scene.
For methods like MVSDet~\cite{xu2024mvsdet} that are not inherently sensor-geometry-free (SG-Free), we adapt them to an SG-Free setting by using predictions (e.g., multi-view poses) from VGGT as input for training and evaluation. Thus, for MVSDet, the time and memory overhead related to the shared VGGT components are included in the analysis. VGGT computation is listed in the first row of the table, while the additional time and memory consumption of each method are shown in subsequent rows.
Our method `w/ Ours' achieve comparable time consumption to `w/ MVSDet' while significantly reducing memory usage. For instance, `w/ Ours' consumes only 3.57 GB of memory compared to 13.81 GB for `w/ MVSDet'. Moreover,  our method achieves a significant 5.6 points performance improvement with 40 input frames.

\noindent{\textbf{Performance Varies with the Number of Frames}}. 
We conduct an ablation study on ScanNet by feeding different numbers of frames into our model during evaluation, as shown in the~\tabref{subtab:frames}. The results demonstrate that more frames provide richer information, enabling the model to achieve higher performance. Performance saturates around 80 frames, suggesting that 80 frames are sufficient to capture most of the information needed to represent a 3D scene.

\begin{table*}[h!]
\centering

\begingroup
{\footnotesize\relsize{+0.3}
\setlength{\tabcolsep}{3pt}        %
\renewcommand{\arraystretch}{0.98} %

\begin{tabularx}{\textwidth}{l|*{19}{>{\centering\arraybackslash}X}}
\toprule
\textbf{Methods} & cab & bed & chair & sofa & tabl & sink & washer & toil & bath & ovn & refg & shelf & stove  & tvm & stl & fire & dshw & \multicolumn{2}{c} {\textbf{mAP@0.25}} \\
\midrule
ImVoxelNet  & 06.3 & 42.4 & 04.4 & 26.7 & 03.4 & 05.9 & 23.5 & 20.9 & 47.1 & 09.1 & 15.0 & 04.4 & 00.0  & 00.0 & 00.1 & 0.00 & 00.8 & \multicolumn{2}{c}{12.4}  \\
NeRF-Det  & 13.3 & 51.7 & 08.8 & 28.1 & 08.2 & 04.5 & 35.7 & 42.4 & 71.7 & 14.8 & 19.4 & 06.6 & 00.1  & 00.0 & 01.3 & 0.00 & 00.2 & \multicolumn{2}{c}{18.1} \\
MVSDet  & 10.9 & 53.0 & 09.0 & 34.3 & 07.7 & 07.4 & 40.1 & 43.8 & 69.8 & 16.3 & 25.5 & 05.7 & 00.5  & 00.0 & 01.2 & 0.00 & 03.7 & \multicolumn{2}{c}{19.4} \\
\rowcolor[gray]{.9} VGGT-Det & 14.6 & 58.8 & 15.2 & 48.2 & 26.3 & 09.6 & 56.4 & 55.1 & 64.9 & 30.9 & 57.7 & 09.1 & 00.0  & 00.0 & 10.9 & 12.5 & 05.9 & \multicolumn{2}{c} {\textbf{28.0~\footnotesize{\textcolor{darkgreen}{(+8.6)}}}}\\
\bottomrule
\end{tabularx}}
\endgroup
\vspace{+0.1cm}
\begingroup
\captionsetup{aboveskip=2pt,belowskip=0pt}
\captionof{table}{%
Comparison of mAP@0.25 on ARKitScenes. To achieve Sensor-Geometry-Free~(SG-Free) setting and ensure a fair comparison, ImVoxelNet~\citep{rukhovich2022imvoxelnet}, NeRF-Det~\citep{xu2023nerf} and MVSDet~\citep{xu2024mvsdet} are retrained with multi-view camera poses predicted by VGGT. Methods lacking both results and code support for ARKitScenes~(e.g., FCAF3D) are not included. Our method significantly outperforms the competitive methods.}
\label{tab:arkit_result}
\vspace{-0.4cm}
\endgroup
\end{table*}

\noindent{\textbf{Performance under Different Feature Aggregation Methods}}.
We analyze the performance of our Query-Driven Feature Aggregation by comparing it with alternative feature aggregation strategies. The results are summarized in~\tabref{subtab:aggregation}.
We consider three feature aggregation methods. 
(i) In the `Vanilla-v1' method, only the encoded features from the last layer of the VGGT encoder~\cite{wang2025vggt} are fed into the decoder. (ii) In the `Vanilla-v2' method, the encoded features from four specific layers of the VGGT encoder—namely, the 4th, 11th, 17th, and 23rd layers (following VGGT~\cite{wang2025vggt})—are sequentially fed into the decoder. 
(iii) Our Query-Driven Feature Aggregation~(`QD' in~\tabref{subtab:aggregation}) employs a See-Query to adaptively interact with object queries across decoder layers, thereby determining and aggregating the most relevant encoded geometric features.
As shown in~\tabref{subtab:aggregation}, `Vanilla-v2' achieves better performance than `Vanilla-v1', demonstrating that incorporating multi-level features provides richer information. Furthermore, our Query-Driven Feature Aggregation outperforms `Vanilla-v2' by \textbf{2.7} points.
This improvement highlights the effectiveness of our design.

By interacting with object queries, the See-Query learns to adaptively aggregate the most relevant geometric features, leading to significant performance gains.

\subsection{Comparison with Alternatives}
\textbf{Quantitative Comparison}.
\tabref{tab:main_result_0.25} presents the quantitative comparison of mAP@0.25 across different methods on the ScanNet dataset~\cite{dai2017scannet}. Notably, our proposed VGGT-Det achieves significant performance improvements over the baseline and the competitive multi-view indoor 3D object detection methods~\citep{imvotenet,xu2023nerf,xu2024mvsdet}.
Specifically, by progressively incorporating the Attention-Guided Query Generation (AG) and Query-Driven Feature Aggregation (QD), our method demonstrates a consistent upward trend in overall mAP, culminating in a final result of \textbf{46.9}, which is \textbf{+4.4 higher} than MVSDet~\citep{xu2024mvsdet}. This highlights the effectiveness of our
method.
Note that to adapt the competitive methods to the same SG-Free setting for a fair comparison,
we use the predicted multi-view poses from VGGT to retrain the models of ImVoxelNet, NeRF-Det and MVSDet.
To provide a more comprehensive analysis, we additionally compare with a representative point cloud–based method FCAF3D~\cite{rukhovich2022fcaf3d}.
To adapt it to the same SG-Free setting for a fair comparison,
we use the predicted point clouds from VGGT~\cite{wang2025vggt} to retrain the models of FCAF3D.
VGGT-Det significantly outperforms FCAF3D by 6.3 points.
In summary, the superior performance of VGGT-Det validates the proposed design's ability to effectively 
 leverage VGGT’s internally learned priors.

We also compare against competitive multi-view indoor 3D object detectors~\cite{xu2023nerf,xu2024mvsdet} on ARKitScenes~\cite{arkitscenes}, as shown in \tabref{tab:arkit_result}. VGGT-Det surpasses the state-of-the-art MVSDet~\cite{xu2024mvsdet} by \textbf{8.6} points, further demonstrating the superiority of our method.
We also find that objects in a few categories, e.g. TV, are typically small, thin, or embedded within cabinetry or walls. Without accurate sensor geometry as prior information, such instances are highly prone to localization errors, leading to universally low performance across all the existing models.
Note that to adapt the competitive methods to the same SG-Free setting for a fair comparison,
we use the predicted multi-view poses from VGGT to retrain the models of ImVoxelNet, NeRF-Det and MVSDet.  

\noindent{\textbf{Qualitative Comparison}}.
The qualitative comparison results are shown in Fig.~\ref{fig:demo_comparison}.
These
results highlight the clear advantages of our VGGT-Det: it consistently detects more
objects than the state-of-the-art
MVSDet~\cite{xu2024mvsdet} and achieves higher detection
precision, benefiting from the effective utilization of
internal VGGT priors by our AG and QD modules.

\vspace{-0.1cm}
\section{Conclusion}
In this work, we introduce a more practical setting: Sensor-Geometry-Free 3D object detection, which removes the dependency on sensor-derived geometric inputs.
To tackle this setting, we leverage the VGGT encoder to build a transformer-based backbone. Then, we propose two key innovations: Attention-Guided Query Generation and Query-Driven Feature Aggregation. The former utilizes the internal semantic priors from VGGT's attention mechanisms to generate object queries that effectively focus on object regions while maintaining the global spatial structure.
The latter adaptively aggregates multi-level geometric
features by  dynamically interacting with object queries.
Extensive experiments show that our VGGT-Det significantly outperforms strong methods in the SG-Free setting on both ScanNet and ARKitScenes. 
Ablations confirm that our AG and QD  effectively leverage the learned priors from pretrained~VGGT.

{
    \small
    \bibliographystyle{ieeenat_fullname}
    \bibliography{main}
}

\clearpage
\setcounter{page}{1}
\maketitlesupplementary

\section{More Ablation Studies}

\noindent{\textbf{{Performance under different $\lambda_{\text{dist}}$}}}.
In this subsection, we study the impact of varying $\lambda_{\text{dist}}$ in~\equref{equ:lambda} on the performance of our proposed Attention-Guided Query Generation (AG). 
As shown in \tabref{tab:ablation_lambda},
the best performance in our evaluation is obtained around  $\lambda_{\text{dist}}=0.8$.
Intuitively, 
 $\lambda_{\text{dist}}$ modulates the balance between attention guidance and spatial dispersion. When $\lambda_{\text{dist}}$
 is too small, the model relies heavily on attention guidance, leading to excessive focus on specific high-attention regions while neglecting the global spatial structure. When $\lambda_{\text{dist}}$
 is too large, the spatial dispersion term dominates, reducing the influence of attention guidance.
 In summary, $\lambda_{\text{dist}}$
 controls an intuitive balance between attention guidance and spatial dispersion. The observed optimal performance around $\lambda_{\text{dist}}=0.8$ in our ablations
 aligns well with our goal of initializing queries that focus on semantic regions while preserving the 
global spatial structure.

\begin{table}[!h]
    \centering
    \resizebox{0.55\linewidth}{!}{
            \begin{tabular}{l|c}
                \toprule
                \textbf{Methods} & \textbf{mAP@0.25}  \\
                \midrule
                w/ AG ($\lambda_{\text{dist}}$=0.9) & 44.9  \\
                w/ AG ($\lambda_{\text{dist}}$=0.8) & 46.9   \\
                w/ AG ($\lambda_{\text{dist}}$=0.5) & 40.7   \\
                \bottomrule
            \end{tabular}
}
\captionsetup{aboveskip=2pt}\captionsetup{belowskip=0pt}\captionof{table}{Performance under different $\lambda_{\text{dist}}$ in~\equref{equ:lambda}.}
\label{tab:ablation_lambda}
\end{table}

\noindent{\textbf{{The impact of varying levels of noise}}}.
In this subsection, we evaluate the robustness of our method to noise in VGGT-predicted point clouds. Because the quality of the pretrained VGGT~\cite{wang2025vggt} outputs is not directly controllable, we introduce controlled noise by adding Gaussian noise to the VGGT point clouds.
Specifically, the noise is sampled from a normal distribution $n \sim \mathcal{N}(0, \sigma^2)$,
where $\sigma$ represents the standard deviation of the noise. To ensure precise and scalable noise
intensities, $\sigma$ is defined as: $\sigma = \mathrm{range} \cdot \text{noise\_level}$,
where $\mathrm{range} = \max(P) - \min(P)$ represents the value range of the point clouds, and $\text{noise\_level} \in [0, 1]$ is a user-defined parameter controlling the noise intensity.
The noisy point cloud is generated as: $P' = P + N$, where $N$ is the noise matrix.
This method enables fine-grained noise control, facilitating precise evaluation of noise effects.
We evaluate VGGT-Det on ScanNet using noisy point clouds. To provide a fair comparison, we also evaluate the representative point cloud–based method FCAF3D~\cite{rukhovich2022fcaf3d}, which likewise takes noisy point clouds as input. The results are summarized in~\tabref{tab:noise_robustness}.
As expected, the performance of both
methods degrades as the
noise level increases, confirming the adverse impact of noise on detection.
However, FCAF3D starts
to degrade significantly at a noise level of 0.001 and drops to 0.0 mAP@0.25 at 0.1,
whereas VGGT-Det
maintains robustness and only starts degrading at 0.1, achieving 34.1 mAP@0.25
even at 0.3.
At the same noise level of 0.01, VGGT-Det significantly outperforms FCAF3D by \textbf{28.3 points}
(47.0 vs. 18.7), demonstrating its superior robustness.

This robustness stems from our Attention-Guided Query Generation, which efficiently leverages VGGT’s internal
semantic priors to generate object queries that focus on semantically relevant regions, as illustrated
in~\figref{fig:attention}. By reducing reliance on the precise geometric details of the predicted point clouds from VGGT,
this mechanism effectively mitigates the impact of noise. In contrast, FCAF3D relies
heavily on the geometric integrity of the input point clouds to extract features, making it significantly
more susceptible to noise, as evidenced by its rapid performance degradation
in~\tabref{tab:noise_robustness}.

\begin{table}[!t]
    \centering
    \resizebox{0.75\linewidth}{!}{
\begin{tabular}{l|c|c}
\toprule
\textbf{Noise Level} & \textbf{FCAF3D} & \textbf{VGGT-Det (Ours)} \\
\midrule
0.30  & 0.0  & 34.1 \\
0.20  & 0.0  & 39.4 \\
0.10  & 0.0  & 44.5 \\
0.01  & 18.7 & 47.0 \\
0.005 & 33.4 & 47.0 \\
0.001 & 39.7 & 46.9 \\
0.000 & 40.6 & 46.9 \\
\bottomrule
\end{tabular}
}
\captionsetup{aboveskip=2pt}\captionsetup{belowskip=0pt}\captionof{table}{Robustness under different noise levels: comparison between FCAF3D~\cite{rukhovich2022fcaf3d} and our VGGT-Det.}
\label{tab:noise_robustness}
\end{table}

\noindent{\textbf{{Performance across different numbers of input frames}}}.
For a fair comparison, we follow the standard practice in indoor multi‑view 3D object detection (e.g., the official MVSDet implementation) by using 80 input frames. Nevertheless, our approach is not restricted to this configuration. Across different numbers of input frames, it consistently surpasses the strongest competing method, MVSDet, as shown in~\tabref{tab:num_frames_supp}.

\begin{table}[htbp]
\centering
\footnotesize %
\begin{tabular}{l|c|c|c|c|c}
\toprule
\textbf{\# of Frames} & \textbf{20} & \textbf{40} & \textbf{60} & \textbf{80} & \textbf{100} \\
\midrule
MVSDet~\cite{xu2024mvsdet}  &
35.9 &
39.7 &
40.9 &
42.5 &
42.5 \\
\textbf{Ours} &
\textbf{42.6} &
\textbf{45.3} &
\textbf{46.2} &
\textbf{46.9} &
\textbf{47.3} \\

\bottomrule
\end{tabular}

 \vspace{-0.3cm}
\caption{Performance across different numbers of input frames}

\label{tab:num_frames_supp}

\end{table}

\noindent{\textbf{{Performance gap to methods that utilize sensor geometry}}}.
In~\tabref{tab:efficiency_full}, M1, M2, and M3 correspond to ImVoxelNet~\cite{imvotenet}, NeRF‑Det~\cite{xu2023nerf}, and MVSDet~\cite{xu2024mvsdet}, respectively, while “+SG” indicates methods that leverage sensor geometry. Our method achieves the highest performance in both settings—SG‑Free and SG‑Based. A noticeable gap remains between the two, which is expected since sensor‑provided geometric priors offer strong cues for detection. Nevertheless, we regard the SG‑Free setting as an important and practical regime, given that sensor geometry is often unavailable or impractical in real‑world scenarios.

\begin{table}[!h]
    \centering
    \resizebox{0.85\linewidth}{!}{
\begin{tabular}{l|c|c|c|c}
\toprule
\textbf{Method} & \textbf{ M1} & \textbf{ M2} & \textbf{ M3} & \textbf{ Ours} \\
\midrule
mAP &  35.2 & 41.2 &  42.5 &  \textbf{46.9} \\
\midrule

\textbf{Method} & \textbf{M1+SG} & \textbf{M2+SG} & \textbf{M3+SG} & \textbf{Ours+SG} \\
\midrule
mAP & 48.1  &49.5  & 56.2  & \textbf{58.8}  \\

\bottomrule
\end{tabular}
}
 \vspace{-0.3cm}
\caption{Performance gap to methods that leverage sensor geometry.}
\label{tab:efficiency_full}

\end{table}

\noindent{\textbf{{More methods in efficiency analysis}}}.
To ensure fair comparison under the same SG‑Free setting, we train and evaluate SG‑Free variants of all models, rather than their original sensor‑geometry‑dependent versions. Specifically, multi‑view RGB images are first processed by VGGT to estimate camera poses, which are then provided to each detector. Consequently, the SG‑Free pipeline inherently includes VGGT’s runtime and memory costs, ensuring fairness in comparison. 
In~\tabref{tab:efficiency_full}, M1, M2, and M3 represent ImVoxelNet~\cite{imvotenet}, NeRF‑Det~\cite{xu2023nerf}, and MVSDet~\cite{xu2024mvsdet}, respectively. 
As shown, our designs (+Ours) achieve inference time comparable to the strongest competitor (+MVSDet), while attaining the lowest GPU memory usage among all methods.

\begin{table}[!h]
    \centering
    \resizebox{0.85\linewidth}{!}{
\begin{tabular}{l|c|c|c|c|c}
\toprule
\textbf{Method} & \textbf{VGGT} & \textbf{+M1} & \textbf{+M2} & \textbf{+M3} & \textbf{+Ours} \\
\midrule
T~(s)  & 0.68 & 0.09 & \textbf{0.07} & 0.21 & 0.23 \\
 M~(GB) & 11.71 & 12.95 & 6.76 & 13.81 & \textbf{3.57} \\

\bottomrule
\end{tabular}
}
 \vspace{-0.3cm}
\caption{Efficiency comparison with different mothods in SG-Free setting with 40 input frames.}
\label{tab:efficiency_full}

\end{table}

\noindent{\textbf{{Impact of view number on efficiency}}}.
As shown in~\tabref{tab:efficiency_different_input_numbers}, the time and GPU memory cost increase with more views, which is a common challenge in multi-view 3D detection.

\begin{table}[!h]
    \centering
    \resizebox{0.85\linewidth}{!}{
\begin{tabular}{l|c|c|c|c|c} %
\toprule
\textbf{\# of Frames} & \textbf{20} & \textbf{40} & \textbf{60} & \textbf{80} & \textbf{100} \\
\midrule
T (s)  & 0.49 & 0.91 & 1.48 & 2.20 & 3.06 \\
 M (GB) & 14.46 & 15.28 & 17.50 & 19.48 & 21.66 \\
\bottomrule
\end{tabular}
}
 \vspace{-0.3cm}
\caption{Efficiency comparison under different input numbers.}
\label{tab:efficiency_different_input_numbers}

\end{table}

\noindent{\textbf{{Visualization analysis}}}.
To further study the effectiveness of our Attention-Guided Query Generation, we conduct a visualization analysis of both the attention maps and the positions of the generated object queries.
As shown in~\figref{fig:attention}, the 2nd column illustrates the attention maps, which clearly highlight the object regions. This observation supports our motivation: attention can provide semantic guidance for query generation.
In the 3rd column, we compare the generated object query positions. The red points represent the object queries generated without attention guidance, while the green points correspond to the queries generated by our Attention-Guided Query Generation. Notably, within the object regions (highlighted by green boxes), our method generates significantly more object queries (green points) compared to the baseline (red points). This demonstrates that our Attention-Guided Query Generation effectively focuses on object regions, leading to a more accurate query distribution.
The effectiveness of our approach is further reflected in an improvement of \textbf{2.8 points} in mAP@0.25, as reported in Tab.~2a of the main paper. This result highlights the direct contribution of Attention-Guided Query Generation to the overall performance.

\section{Comparison with Alternatives}

\noindent{\textbf{{Qualitative Comparison}}}.
We provide a qualitative comparison with the best-performing competitive method, MVSDet~\cite{xu2024mvsdet}. To adapt to the
Sensor-Geometry-Free~(SG-Free) setting and ensure a fair comparison,
MVSDet is trained and tested using multi-view poses predicted by VGGT~\cite{wang2025vggt}. 
As shown in \figref{fig:cmp_samples}, our VGGT-Det detects more objects with higher accuracy, which is consistent with the significant performance improvement of \textbf{4.4 points} reported in Tab.~1 of the main paper.
The higher performance benefits from the effective utilization of internal VGGT priors by the proposed Attention-Guided Query Generation~(AG) and Query-Driven Feature Aggregation~(QD) modules.

\section{Training and Testing Time}
All the ablation experiments are conducted on eight H800 GPUs. Training our model on the ScanNet dataset~\cite{dai2017scannet} takes approximately 2 days to complete. For testing, the entire ScanNet testing set can be processed in about 1 minute. 

\section{Limitation and Future work}
While VGGT-Det achieves significant improvements over strong alternative methods, several limitations remain for further exploration. Across current SG-free pipelines, VGGT incurs noticeable runtime and memory overhead. Besides, because VGGT produces normalized predictions, the scales from datasets
are utilized to denormalize predictions of VGGT in all  the current SG-free pipelines. Looking ahead, introducing a lighter VGGT-like model with metric-scale predictions could further advance this direction.

\begin{figure*}[h!]
     \centering
    \begin{overpic}[width=1\textwidth]
    {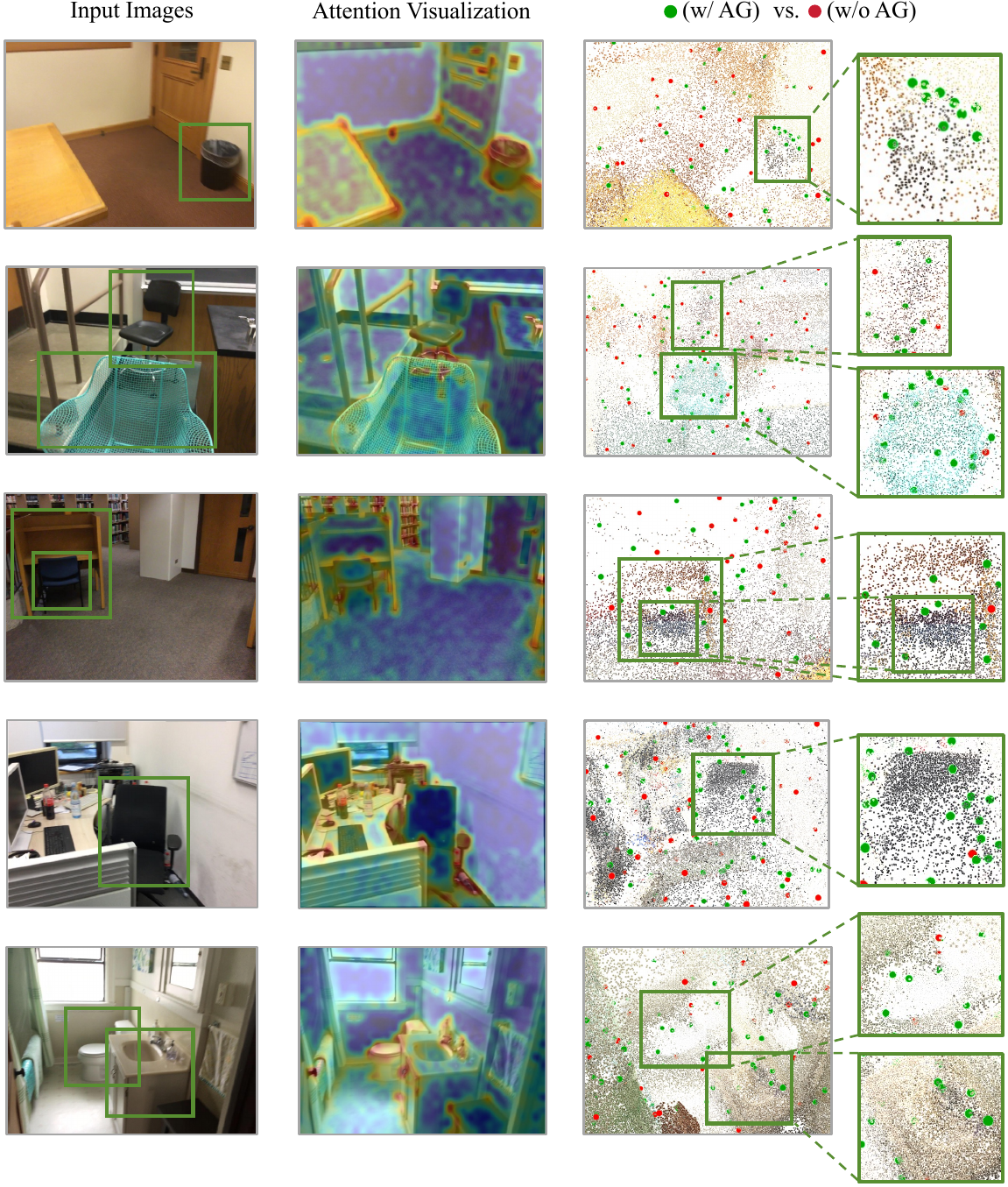}
    \end{overpic}
    \vspace{-0.2cm}     
    \caption{Visualization of attention and generated object query positions. Compared to farthest point sampling without guidance~(red points), the points sampled by AG~(green points) are more concentrated in object regions~(labeled by green boxes),
resulting in more green points than red points in those areas. For clarity, we recommend viewing the figure in color and zooming in.}
    \label{fig:attention}
\end{figure*}

\begin{figure*}[h!]
     \centering
    \begin{overpic}[width=1\textwidth]
    {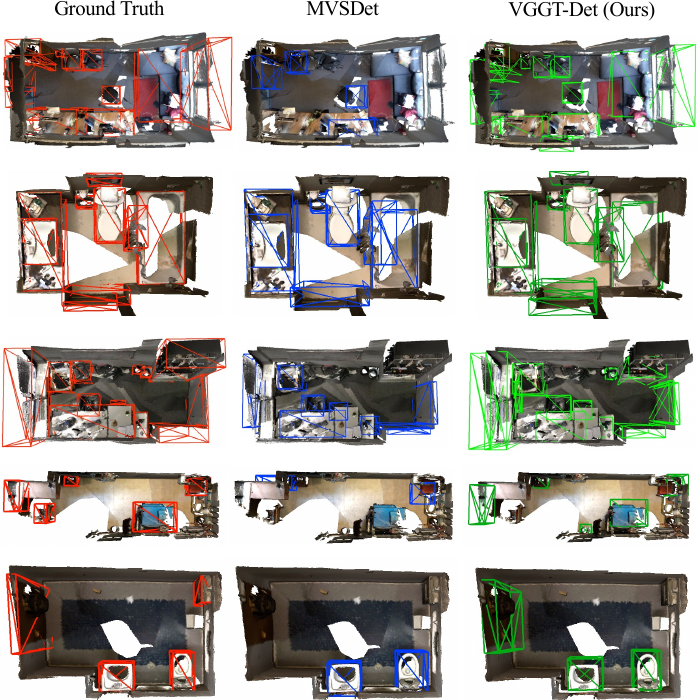}
    \end{overpic}
    \vspace{-0.2cm}     
    \caption{Qualitative comparison with MVSDet~\cite{xu2024mvsdet}. To achieve the Sensor-Geometry-Free~(SG-Free) setting and ensure a fair comparison,
MVSDet is trained with multi-view poses predicted by VGGT~\cite{wang2025vggt}.
The mesh here is not utilized
in the methods and is only for visualization.}
    \label{fig:cmp_samples}
\end{figure*}

\end{document}